\documentclass[conference]{IEEEtran}
\IEEEoverridecommandlockouts
\usepackage{cite}
\usepackage{amsmath,amssymb,amsfonts}
\usepackage{algorithmic}
\usepackage{graphicx}
\usepackage{textcomp}
\usepackage{xcolor}
\usepackage{adjustbox}
\usepackage{hyperref}
\usepackage{caption}
\usepackage{natbib}
\usepackage{CJKutf8}

\def\BibTeX{{\rm B\kern-.05em{\sc i\kern-.025em b}\kern-.08em
    T\kern-.1667em\lower.7ex\hbox{E}\kern-.125emX}}
\begin{document}

\begin{CJK*}{UTF8}{gbsn}

\title{Exploring the Capabilities of ChatGPT in Ancient Chinese Translation and Person Name Recognition
}

\author{\IEEEauthorblockN{Shijing Si$^{\dag}$}
\IEEEauthorblockA{\textit{School of Economics and Finance} \\
\textit{Shanghai International Studies University}\\
Shanghai, China \\
shijing.si@shisu.edu.cn}
\and
\IEEEauthorblockN{Siqing Zhou$^{\dag}$\thanks{$^{\dag}$:Shijing Si and Siqing Zhou are co-first authors.}}
\IEEEauthorblockA{\textit{School of Economics and Finance} \\
\textit{Shanghai International Studies University}\\
Shanghai, China \\
zsq020104@gmail.com}
\and
\IEEEauthorblockN{Le Tang}
\IEEEauthorblockA{\textit{Library} \\
\textit{Shanghai International Studies University}\\
Shanghai, China \\
1067061406@qq.com}
\and
\IEEEauthorblockN{Xiaoqing Cheng}
\IEEEauthorblockA{\textit{School of Mathematics and Statistics
} \\
\textit{Xi'An Jiaotong University}\\
Shaanxi, China \\
xiaoqing9054@mail.xjtu.edu.cn}
\and
\IEEEauthorblockN{Yugui Zhang}
\IEEEauthorblockA{
\textit{School of Economics and Finance}\\
\textit{Shanghai International Studies school} \\
Shanghai, China \\
yuguizhang@126.com}
}

\maketitle

\begin{abstract}
 ChatGPT's proficiency in handling modern standard languages suggests potential for its use in understanding ancient Chinese. 
 This paper explores ChatGPT's capabilities on ancient Chinese via two tasks: translating ancient Chinese to modern Chinese and recognizing ancient Chinese names. 
 A comparison of ChatGPT's output with human translations serves to evaluate its comprehension of ancient Chinese. The findings indicate that: (1.)the proficiency of ancient Chinese by ChatGPT is yet to reach a satisfactory level; (2.) ChatGPT performs the best on ancient-to-modern translation when feeding with three context sentences.
 To help reproduce our work, we display the python code snippets used in this study.
\end{abstract}


\section{Introduction}

Ancient languages serve as repositories for humanity's historical and cultural heritage. 
 As a treasure and important heritage of ancient Chinese culture \citep{lee2012glimpses}, ancient Chinese has important cultural significance and historical value in the study of natural language processing.

Recently there have been considerable advancements made in the analysis and interpretation of ancient Chinese through the use of deep learning models
\citep{cheng2020integration,tian2021anchibert,tang2022simple,anderson2023proceedings,wang2023enhancing}.
However, the understanding of ancient Chinese is a very challenging task in natural language processing due to its complex grammatical structures, 
cultural nuances, and polysemy of the 
language \citep{yu2019machine,jin2023morphological,guo2023towards}. 

Deep learning methods
has achieved significant progress in 
many natural language processing (NLP) tasks
\citep{si2019sentence,devlin2018bert,si2020students,brown2020language}.
Notably, large language models like ChatGPT
have been shown powerful generation and
understanding capabilities across many languages \citep{sundararaman2020methods,lai2023chatgpt,liu2023chatgpt,ray2023chatgpt,bian2023chatgpt},
and has attracted wide attention and application around the world since its release \citep{fang2023chatgpt,wang2023chatgpt,wu2023brief,kocon2023chatgpt,gilardi2023chatgpt,ronanki2023chatgpt}.

However, there are few research on the use of ChatGPT for ancient Chinese processing. \citet{jin2023morphological}
evaluates the performance of a few machine translation methods including ChatGPT. \cite{lan2023chatgpt} systematically examined ChatGPT at six cognitive levels on the Book of changes (易经, \textit{Yi Jing} in Chinese).
Following these works, we study the capability of ChatGPT on 
ancient-to-modern Chinese translation and people 
name recognition \citep{zhang2021people} in ancient Chinese text.
We leveraged the ancient Chinese book, called \textit{Shi Shuo Xin Yu}(世说新语 in Chinese characters), to assess ChatGPT's ability in understanding ancient Chinese through two tasks: the task of translating ancient Chinese into modern Chinese and the task of recognizing people's names in ancient Chinese.

In this paper, we evaluate the capability of ChatGPT on
an ancient Chinese book, \textit{Shi Shuo Xin Yu}, which is largely ignored by previous research. Also we study
the performance of ChatGPT on ancient-to-modern translation by varying the input length for each individual query. Additionally, the personal name recognition is rarely explored in the ancient Chinese processing. To summarize, our cntributions are shown as follows:
\begin{itemize}
    \item We evaluate the ancient-to-modern Chinese translation of ChatGPT on \textit{Shi Shuo Xin Yu}, and observe that the translation quality depends on the context length. From our experiments, ChatGPT performs the best on ancient-to-modern translation when three consecutive sentences are fed into the prompt instruction.
    \item We also investigate the capability of ChatGPT on ancient person name recognition, finding that ChatGPT outperforms the commonly used method in Jieba, and its performance improves as the number of demonstrations increases.
\end{itemize}

The article is organized as follows. Section \ref{sec:rel} presents the related works on ChatGPT and ancient Chinese processing.
Section \ref{sec:exp} demonstrates the
experimental setup and implementation code. Section \ref{sec:res}
exhibits the experimental results and analysis. Section \ref{sec:conc} shows the conclusion and discussion.

\section{Related Works}\label{sec:rel}

ChatGPT and other large language models have indeed garnered significant attention due to their impressive capabilities \cite{Ciesla24}.
However, these large model are not perfect and still have limitations, including producing inconsistent or nonsensical responses, or writing information that seems factual but is actually completely made up. They are sensitive to the way prompts are phrased, and fail to possess any real understanding of the world or self-awareness.
Therefore, there are many studies that explore and evaluate the performance of ChatGPT on many tasks and examinations. We will
illustrate related works in three subsections.

\subsection{Application of ChatGPT in Natural Language Processing}

\citet{ogundare2023comparative} utilized spontaneous quality (SQ) scores to compare the performance of ChatGPT on many NLP tasks such as machine translation, machine summarization, question answering, and language generation, compared with other mainstream algorithms. \citet{wang2023documentlevel}, taking document-level machine translation (MT) as an experimental platform, conducts an in-depth evaluation of ChatGPT's discourse modeling ability in three aspects: the role of discourse awareness cue, the comparison of translation models, and the analysis of discourse modeling ability. \citet{zhang2023crosslingual} explores ChatGPT's ability to summarize and evaluate on cross-lingual cross-time summarization (CLCTS) tasks. \citet{ronanki2023chatgpt} explores the application of ChatGPT for user story quality assessment and compares its performance to existing benchmarks. \citet{sun2023pushing} proposed a set of generic modules, attempting to break through the limitations of ChatGPT on various NLP tasks, including question answering, common sense reasoning, natural language reasoning, etc. \citet{antoun2023towards} proposed a method for developing and evaluating ChatGPT detectors for French text, focusing on their robustness against extra-territorially based data and common attack schemes. \citet{wang2023chatgpt} explore whether ChatGPT can be a cost-effective supplement to expert feedback by acting as an automated teacher or coach.

\subsection{Application of ChatGPT on Other Fields}

Many research explore the ability of ChatGPT to understand texts in other fields than NLP. \citet{cao2023assessing} investigated ChatGPT's potential cultural background by analyzing its answers to questions aimed at quantifying human cultural differences. The results of the study showed that when inspired by an American context, ChatGPT showed strong consistency with American culture, but it was less effective in adapting to other cultural contexts. \citet{george2023review} explore how ChatGPT can enhance e-commerce through chat and other sectors such as education, entertainment, finance, health, news, and productivity, and how ChatGPT can be used to create more personalized content for users and help businesses improve the efficiency and effectiveness of customer service. \citet{rezayi2023exploring} explored a new field of agricultural natural language processing by studying the effectiveness of pre-training transformer-based language models using food-related text corpora. \citet{wang2023assessing} investigated ChatGPT's ability to infer dynamic network structures from temporal text data, especially financial news. \citet{khondaker2023gptaraeval} conducted a large-scale evaluation of ChatGPT on a wide range of Arabic NLP tasks.  \citet{wu2023qualifying} evaluated ChatGPT's ability in knowledge in the medical field on the Chinese National Medical Licensing Examination (CNMLE).

\subsection{Application of Large Language Models in Ancient Chinese}

However, there is relatively little research on ChatGPT's understanding of ancient Chinese. The research on the understanding of ancient Chinese by language model is mainly based on BERT model. \citet{wang2022SikuBERT} and \citet{chang2023sikugpt} built a pre-trained language model of SikuBERT and SikuGPT for ancient text intelligent processing tasks based on BERT deep language model framework using the verified high-quality full-text corpus of SikukuQuanshu as a training set. Some researchers built GuwenBERT based on RoBERTa model trained on a large number of ancient literature corpus. Tsinghua University has constructed BERT CCPoem, a pre-training model based on BERT specifically for ancient Chinese poetry, which is trained on the complete collection of ancient Chinese poetry CCPC full v1.0 and can be used for intelligent poetry retrieval, recommendation and sentiment analysis.
\citet{wang2023gujibert} introduced two language models, GujiBERT and GujiGPT, which are foundational 
models specifically designed for intelligent information processing of ancient texts.

\section{Experiments}\label{sec:exp}
This section details 
the experimental setup, including the datasets,
and python code for calling ChatGPT API, and
results.

\subsection{Data preparation}

The main data comes from Github open source project NiuTrans/classic-Modern\footnote{The resource can be found in website:\url{https://github.com/NiuTrans/Classical-Modern}}, which is a very comprehensive ancient Chinese-modern parallel corpus, including a large number of classical ancient works. This dataset divides and displays each ancient book by chapters, and the text is stored in the txt documents under each chapter. These bilingual data are provided in sentence level, and three data formats are provided: original text, translation and bilingual.

In this study, the team mainly utilized the bilingual data of the book, \textit{Shi Shuo Xin Yu}, including 36 chapters, to satisfy our needs in the experiments. And most articles in this book are about biographies of famous people in ancient China, which can be used for ancient-to-modern Chinese translation and person name recognition.

Due to the input length limitation of ChatGPT, we split the whole book into 3923 sentences and each time we feed the model a sentence.
For the ancient-to-modern Chinese translation task, we include all sentences, but for the ancient Chinese name recognition task, we randomly select a total of 300 sentences because of time constraint.
Also we manually label Chinese names for the 300 sentences. The data is shown in Table \ref{tab:data} as fowllows.
\begin{table*}[t]
\centering
\caption{\label{tab:data} A few examples used in the experiments. The sentences are extracted from \textit{Shi Shuo Xin Yu}, and Chinese names are manually labeled by the authors.}
\scalebox{1.3}{
\begin{tabular}{ll}
\hline
 Sentence & Names \\ \hline
   孙秀既恨石崇不与绿珠，又憾潘岳昔遇之不以礼。       &  孙秀、石崇、潘岳     \\
     后秀为中书令，岳省内见之，因唤曰：孙令，忆畴昔周旋不？      &  孙秀、潘岳     \\
      岳于是始知必不免。     & 潘岳      \\
   冀神理绵绵常，不与气运俱尽耳！        &  No names     \\\hline
\end{tabular}}
\end{table*}


\subsection{Accesss to ChatGPT}
\label{sect:pdf}

ChatGPT, a pre-trained large language model, simply takes a text input following a \textit{prompt}, then it provides a response to the question. ChatGPT supports HTTP requests in multiple languages. In this experiment, we access ChatGPT via the module \textbf{openai} in Python. There are many pre-trained models available in this module, like gpt-4, gpt-3.5-turbo, gpt-3.5-turbo-0613, etc. In this experiments, we choose gpt-3.5-turbo, as it is the most popular one.

\subsection{Translation of ancient Chinese by ChatGPT}
\label{ssec:trans}

\begin{figure}
    \centering
   \includegraphics[scale=1.6]{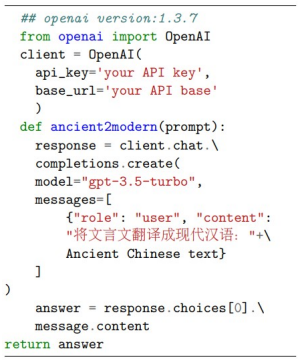}
    \caption{The code snippet for ancient-to-modern Chinese translation. The input of the
chat function is the ancient Chinese sentence,
and the output is its corresponding translation.
} 
\label{fig:trans.code}
\end{figure}

Once having access to ChatGPT in Python, we can simply design a sensible prompt to perform the ancient-to-modern Chinese translation task. We tried a variety of prompt input methods, such as ``Please translate the following ancient Chinese:", ``Translate the ancient Chinese after the colon into modern Chinese:", ``Translate the ancient Chinese into modern Chinese" and so on. We found that it is necessary to include ``translation" and ``modern Chinese" in order to perform the ancient-to-modern Chinese translation robustly, otherwise the output of ChatGPT may have redundant and wrong information.
In addition, we also found that even if we input only ancient Chinese sentences without any instruction, ChatGPT can sometimes return the correct translation results in modern Chinese. For reproduction of our experiments, the python code 
snippet is shown in Fig. \ref{fig:trans.code}.

After many trials, we choose a straightforward ``将文言文翻译成现代汉语(translate ancient Chinese into modern Chinese)" as the prompt instruction for the translation task.
This prompt almost never produces error information, and no superfluous information. we extract the text content that contains only the modern Chinese translation of the ancient Chinese.
Due to the input length constraints set by OpenAI, it is unfeasible to command ChatGPT to translate the entire \textit{Shi Shuo Xin Yu} book directly. Consequently, we use a systematic FOR loop to sequentially translate preloaded segments of the book.

To proportionately evaluate its performance, we adjust the size of the input - the amount of text that needs to be translated.
We start off with a single sentence to observe how well the program can perform a basic translation task. Then, they 
slowly increase the input length to three, five, and eight sentences. This gradual scaling may challenge the program and test its limits, especially in maintaining cohesiveness and continuity in the translation of longer passages.

For the ancient-to-modern translation task, we evaluate
the performance of ChatGPT with
BLEU \citep{papineni2002bleu} and BERT-Score\citep{zhang2019bertscore}.

\subsection{People Name Recognition from Ancient Chinese Literature}

\begin{figure}
    \centering
   \includegraphics[scale=1.6]{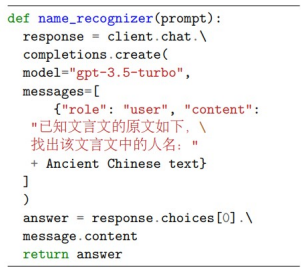}
    \caption{The code snippet for people name
recognition from ancient Chinese text. The
input of the function is the ancient Chinese
sentence, and the output is people name with
analysis.
} 
\label{fig:ner.code}
\end{figure}

Besides translation, we assess the competence of ChatGPT in recognizing names referenced within the ancient Chinese text. It is noteworthy that the effectiveness of ChatGPT in executing this task is contingent upon the prompt instruction.
We utilize a trial-and-error approach to ascertain a suitable prompt instruction,
 "已知文言文的原文如下，找出该文言文中的人名", which means that
 "Given the following ancient Chinese text, find the people names in it". 
For reproduction of our experiments, the python code 
snippet is shown in Fig. \ref{fig:ner.code}.

 The Jieba module in Python is widely employed for word segmentation and rudimentary named entity recognition in Chinese texts. Hence, for our study on the recognition of people names in ancient Chinese text, we incorporate Jieba as a baseline method for comparison.
For the people name recognition task, we evaluate the performance of ChatGPT with commonly used metrics, like precision, recall and F1 score \citep{jiang2016evaluating}.

\section{Results and Analysis}\label{sec:res}
In this section, we exhibit the experimental outcomes and deliver an in-depth analysis.

\subsection{Assessing the Efficacy of ChatGPT in the Translation of Ancient Chinese Texts}
As mentioned in the experimental setup section \ref{sec:exp}, we deploy ChatGPT to convert sentences from the ancient Chinese text, \textit{Shi Shuo Xin Yu}, into modern Chinese in configurations of 1, 3, 5, and 8 sentences per query. These translations are then evaluated and contrasted using quantitative metrics such as BLEU, and BERT-Score. 
Table \ref{tab:blue} illustrates the BLEU scores resulting from the translation of ancient to modern Chinese from ChatGPT under four different configurations: one sentence (1-sent.), three sentences (3-sent.), five sentences (5-sent.), and eight sentences (8-sent.) per individual query.
In this table, we conduct an evaluation of the BLEU scores across three different levels of granularity: 1-gram, 2-gram, and 4-gram.
From Table \ref{tab:blue},
 it is observed that ChatGPT renders the most proficient translations from ancient to modern Chinese when it is engaged with three sequential ancient Chinese sentences (3-sent.), garnering 22.60\% at 2-gram and 20.19‰ at 4-gram Bleu scores. This suggests that a shorter contextual input enhances the translation quality from ancient to modern Chinese. Interestingly, the quality of translation does not exhibit improvement when the context increases to eight sentences, mirroring the performance with a single sentence input. Another important
 finding is that the BLUE scores are still very low, reaching 0 at 4-gram BLUE in all four settings, which means that ChatGPT with GPT-3.5-Turbo still performs very poorly in ancient-to-modern Chinese translation.

\begin{table}[h]
\begin{center}
\caption{\label{tab:blue} The 1-gram, 2-gram, and 4-gram BLUE scores of translated modern Chinese texts from ChatGPT under four settings: one sentence (1-sent.), three sentences (3-sent.), five sentences (5-sent.), and eight sentences (8-sent.) per query. }
\scalebox{1.3}{
\begin{tabular}{c|ccc}
\hline 
&  BLEU &  BLEU &  BLEU \\
&(1-gram)&(2-gram)&(4-gram) \\ \hline
1-sent.& 18.22\% & 13.59‰ & 0  \\
3-sent. & \textbf{22.60\%} & \textbf{20.19‰} & 0  \\
5-sent. & 22.11\% & 12.89‰  & 0 \\
8-sent. & 18.22\% & 13.59‰ & 0\\
\hline
\end{tabular}
}
\end{center}
\end{table}

\begin{table}[h]
\begin{center}
\caption{\label{tab:bert.score}  The BERT-Scores, precision, recall and F1 score, of ancient-to-modern Chinese translation from ChatGPT under four settings: one sentence (1-sent.),
three sentences (3-sent.), five sentences (5-sent.), and eight sentences (8-sent.) per query.}
\scalebox{1.3}{
\begin{tabular}{c|ccc}
\hline 
&  Recall&  Precision &  F1 Score \\ \hline
1-sent.& 76.5\% & 75.9\% & 77.5\% \\
3-sent. & \textbf{77.9\%} & \textbf{77.3\%} & \textbf{78.5\%}  \\
5-sent. & 77.5\% & 77.0\% & 78.0\% \\
8-sent. &77.0\% & 76.0\% & 78.0\%\\
\hline
\end{tabular}}
\end{center}
\end{table}

\begin{figure*}
    \centering
   \includegraphics[width=\textwidth]{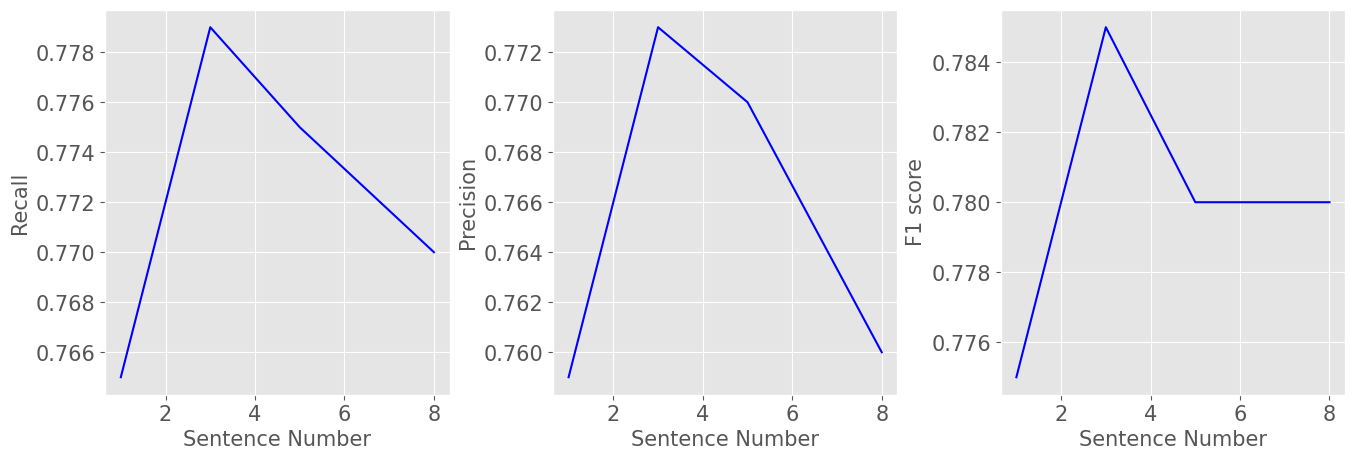}
    \caption{The BERT-scores of ancient-to-modern Chinese translation when ChatGPT is fed with 1, 3, 5, and 8 consecutive sentences.}
    \label{fig:trans.score}
\end{figure*}

Table \ref{tab:bert.score} shares the same structure as Table \ref{tab:blue}, but displays the BERT-Scores, precision, recall, and F1 scores associated with the translation of ancient to modern Chinese by ChatGPT. From this table, ChatGPT performs the best when translating three consecutive ancient Chinese sentences (3-sent.),
achieving 78.5\% F1 score, followed by 5-sent. and 8-sent. This finding is consistent to BLUE scores in Table \ref{tab:blue}.

To visualize the
effects of input sentence number on the translation quality, Fig. \ref{fig:trans.score} presents the line plots of BERT scores versus input sentence number in the prompt.
From this figure, it can be observed that BERT scores (recall, precision and F1 score) peak at 3. This can be explained by the fact that ancient Chinese is complex and context is very important for understanding the pronouns and references. But long context may further confuse the model, so the performance is best at a moderate length of 3 sentences.

\begin{table*}[h]
\centering
\caption{\label{tab:case} The groundtruth (GT) and translated texts from ChatGPT under four settings: one sentence (1-sent.), three sentences (3-sent.), five sentences (5-sent.), and eight sentences (8-sent.) per query.}
\scalebox{1.2}{
\begin{tabular}{c|l}
\hline Method &  Translation  \\ \hline
GT & [CN]:孙秀既怨恨石崇不肯送出绿珠，又不满潘岳从前对自已不礼貌。 \\
& [EN]:Sun Xiu both resented Shi Chong for not \textbf{giving him Zhu Lv} (name of a woman), \\
&  and was dissatisfied with Pan Yue for his past discourteous behavior towards him.\\\hline
1-sent. &[CN]:孙秀既恨石崇不娶绿珠，又遗憾潘岳以前对待他不够礼貌。 \\
& [EN]:Sun Xiu both held resentment towards Shi Chong for not \underline{marrying Zhu}  \\
& \underline{Lv}, and also lamented over Pan Yue's past discourteous behavior towards him. \\\hline
3-sent. & [CN]:孙秀既恨石崇不给绿珠，又懊悔潘岳以前的不礼貌之举。\\
& [EN]:Sun Xiu felt resentment towards Shi Chong for not \textbf{giving the Zhu Lv} \\
&  and regret over Pan Yue's past discourteous actions.\\\hline
5-sent. & [CN]:孙秀既因为石崇未能跟绿珠在一起而怨恨他，又因为潘岳当年未以礼待他而不满。
\\
& [EN]:Sun Xiu both resented Shi Chong for his inability \underline{to be with Zhu Lv} and  \\
&was dissatisfied due to Pan Yue's lack of respect towards him in the past.\\\hline
8-sent. &[CN]:孙秀既憎恨石崇没有娶绿珠为妻，又心怀憾恨潘岳昔日见面时没有礼遇自己。\\
& [EN]:Sun Xiu harbored resentment towards Shi Chong for not \underline{marrying Zhu Lv}, \\
&and also felt regret over Pan Yue's discourtesy towards him in their past encounter.\\
\hline
\end{tabular}}
\end{table*}

To gain deep insights,
we select and analyze a specific sentence as a case study.
The original ancient Chinese sentence is ``孙秀既恨石崇不与绿珠，又憾潘岳昔遇之不以礼。”, which means that
Sun Xiu both resented Shi Chong for not \textbf{giving him Zhu Lv} (name of a woman), and was dissatisfied with Pan Yue for his past discourteous behavior towards him. The groundtruth modern Chinese and translated texts from four settings are exhibited in
Table \ref{tab:case}. To aid readers understand the translated Chinese, we also provide an English version for all methods. 
As shown in Table \ref{tab:case}, 
The translated modern Chinese texts
from ChatGPT is closest to the groundtruth when feeding three sentences (3-sent.). The wrong points are highlighted
by the underlines.

\subsection{Assessing the Efficacy of People Name Recognition in Ancient Chinese using ChatGPT}

People name recognition task can
assess the capability of models on understanding ancient Chinese.
For the people name recognition task, a total of 300 sentences are randomly selected, and a few examples are shown in Table \ref{tab:data}.

\begin{table}[h]
\begin{center}
\caption{\label{tab:nr}Comparative performance of ChatGPT (GPT-3.5-Turbo) and Jieba in recognizing people names from 300 sentences in \textit{Shi Shuo Xin Yu}.}
\scalebox{1.3}{
\begin{tabular}{cccc}
\hline &  Precision &  Recall &F1 score\\ \hline
\multicolumn{4}{c}{Zero-shot performance}\\
Jieba& 44.63\% & 52.94\% & 48.43\% \\
ChatGPT & \textbf{72.66\%} &\textbf{55.11\%} & \textbf{62.68\%} \\
\hline
\multicolumn{4}{c}{ChatGPT Few-shot performance}\\
One-shot & 72.76\% &55.30\% & 62.90\% \\
Five-shot & 72.86\% &55.68\% & 63.12\% \\
Ten-shot & \textbf{73.80\%} &\textbf{56.82\%} & \textbf{64.21\%} \\\hline
\end{tabular}
}
\end{center}

\end{table}

\begin{figure*}
    \centering
   \includegraphics[width=\textwidth]{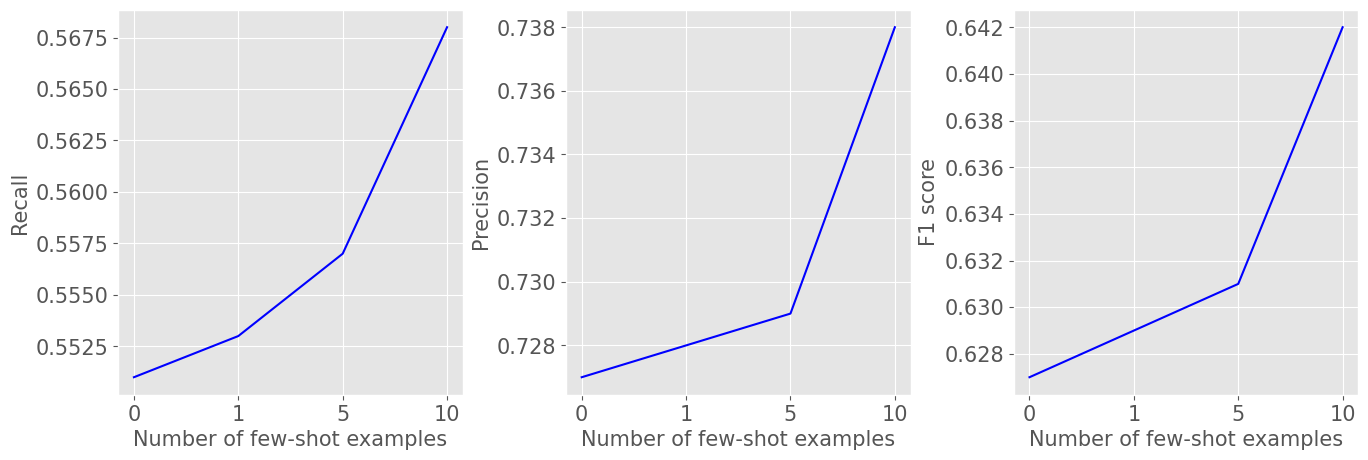}
    \caption{The performance of ChatGPT on person name recognition versus the number of examples in the prompt.}
    \label{fig:name.score}
\end{figure*}
Table \ref{tab:nr} presents the performance of ChatGPT (GPT-3.5-Turbo) and Jieba in recognizing people names from ancient Chinese texts in few-shot settings. This evaluation employed 300 sentences sourced from \textit{Shi Shuo Xin Yu}.
From this table, on the ancient person name recognition task, ChatGPT significantly outperforms Jieba in zero-shot setting, achieving 62.68\% F1 score. But it still has much space to improve.

For the few-shot setting, the performance of ChatGPT gradually improves as the number of demonstrations increases in the prompt.
Fig. \ref{fig:name.score} exhibits this increasing trend as well. As the number of instances increases from 0 to 10, the F1 score rises from 0.628 to 0.642.

\subsection{Limitations}

In our experiments, we did not systematically design and test prompt templates. We just pick the one that performs well on a few examples. In the future, we will investigate how to design efficient prompt templates for in-context learning.

During our experiments, we find that ChatGPT faces many challenges in recognizing personal names from ancient Chinese texts. Fully understanding ancient Chinese sentences rely heavily on the context of the sentence. Sometimes
a sentence may omit the name of the person, or replace it with other names. Therefore, feeding multiple sentences in the prompt of ChatGPT. may enhance the performance as well. We leave it for future research.

\section{Conclusion and Discussion}\label{sec:conc}

Through experiments on the capacity of ChatGPT on ancient-to-modern Chinese translation and person name recognition, we find that there are still much space for improvement. One possible explanation might be that ChatGPT is mainly pre-trained on English corpora with limited amount of ancient Chinese corpora.
While the proficiency of ChatGPT in ancient Chinese remains somewhat restricted, we assert that engagements of deeper research and enhancements - coupled with an abundant corpus of ancient Chinese and specialized knowledge - can ameliorate its proficiency in comprehending and translating and understanding ancient Chinese text. This improvement can potentially expand the possibilities for studying ancient history and culture, along with fostering the preservation and progression of traditional cultural heritage.


\bibliographystyle{IEEEtranN}
\bibliography{IEEEabrv,myref}

\end{CJK*}
\end{document}